\documentclass[conference]{IEEEtran}

\usepackage[numbers]{natbib}
\usepackage{multicol}
\usepackage[bookmarks=true]{hyperref}

\usepackage{graphics}    
\usepackage{times}       
\usepackage{amsmath}     
\usepackage{amssymb}     
\usepackage{graphicx}
\usepackage[noline, ruled]{algorithm2e}
\usepackage[noend]{algpseudocode}
\usepackage{subfigure}
\usepackage{mathtools}
\usepackage[usenames, dvipsnames]{color} 
\usepackage{float}
\usepackage{multirow}
\usepackage[export]{adjustbox} 
\usepackage{booktabs}
\usepackage{caption}
\usepackage{rotating}  
\usepackage{xspace}
\usepackage{pifont}

\let\labelindent\relax 
\usepackage{enumitem}

\newcommand{\cmark}{\ding{51}}%
\usepackage[font=small]{caption}

\def\secref#1{Sec.~\ref{#1}}
\def\figref#1{Fig.~\ref{#1}}
\def\tabref#1{Tab.~\ref{#1}}
\def\eqref#1{Eq.~(\ref{#1})}

\renewcommand{\vec}[1]{\mathbf{#1}}	
\newcommand{\mat}[1]{\mathbf{#1}}  	
\newcommand{\set}[1]{\mathcal{#1}} 	

\newcommand{\RR}{\mathbb{R}} 

\def\argmax{\mathop{\rm argmax}}

\usepackage{array}
\newcolumntype{L}[1]{>{\raggedright\let\newline\\\arraybackslash\hspace{0pt}}m{#1}}
\newcolumntype{C}[1]{>{\centering\let\newline\\\arraybackslash\hspace{0pt}}m{#1}}
\newcolumntype{R}[1]{>{\raggedleft\let\newline\\\arraybackslash\hspace{0pt}}m{#1}}

\makeatletter
\DeclareRobustCommand\onedot{\futurelet\@let@token\@onedot}
\def\@onedot{\ifx\@let@token.\else.\null\fi\xspace}

\def\eg{e.g\onedot} 
\def\ie{i.e\onedot}

\def\etal{\emph{et al}\onedot}
\makeatother

\newif\ifrssfinal
\rssfinalfalse

\newif\ifrssdraft
\rssdraftfalse

\definecolor{purple}{rgb}{0.5, 0.0, 0.5}
\definecolor{orange}{rgb}{0.6, 0.35, 0}
\definecolor{lightgreen}{rgb}{0.68, 1, 0.18}
\definecolor{darkgreen}{rgb}{0.09, 0.32, 0.24}
\definecolor{darkred}{rgb}{0.6, 0, 0}
\definecolor{brown}{rgb}{0.64, 0.16, 0.16}
\definecolor{lilac}{rgb}{0.82, 0.67, 1}

\makeatletter
\newcommand\footnoteref[1]{\protected@xdef\@thefnmark{\ref{#1}}\@footnotemark}
\makeatother

\title{OverlapNet: Loop Closing for LiDAR-based SLAM}

\author{Xieyuanli Chen$^*$ \qquad Thomas L\"abe$^*$ \qquad 
\qquad Andres Milioto$^*$ \qquad Timo R\"ohling$^{*,\ddagger}$\\
Olga Vysotska$^{\dagger,*}$ \qquad Alexandre Haag$^{\dagger}$ \qquad Jens Behley$^*$ \qquad Cyrill Stachniss$^*$\\[2mm]
$^*$Photogrammetry \& Robotics Lab, University of Bonn, Germany\\
$^\ddagger$Fraunhofer FKIE, Wachtberg, Germany\\
$^{\dagger}$Autonomous Intelligent Driving GmbH, Munich, Germany
}

\begin{document}
\maketitle

\IEEEpeerreviewmaketitle
\thispagestyle{empty}
\pagestyle{empty}

\begin{abstract}
  Simultaneous localization and mapping (SLAM) is a fundamental capability required by most autonomous systems.
  In this paper, we address the problem of loop closing for SLAM based on 3D laser scans recorded by autonomous cars.
  Our approach utilizes a deep neural network exploiting different cues generated from LiDAR data for finding loop closures. It estimates an image overlap generalized to range images and provides a relative yaw angle estimate between pairs of scans.
  Based on such predictions, we tackle loop closure detection and integrate our approach into an existing SLAM system to improve its mapping results.
  We evaluate our approach on sequences of the KITTI odometry benchmark and the Ford campus dataset.
  We show that our method can effectively detect loop closures surpassing the detection performance of state-of-the-art methods.
  To highlight the generalization capabilities of our approach, we evaluate our model on the Ford campus dataset while using only KITTI for training. The experiments show that the learned representation is able to provide reliable loop closure candidates, also in unseen environments.
\end{abstract}

\section{Introduction}
\label{sec:intro}

Simultaneous localization and mapping or SLAM~\cite{bailey2006ram1,stachniss2016handbook-slamchapter} is an integral part of most robots and autonomous cars. Graph-based SLAM  often relies on (i)~pose estimation relative to a recent history, which is called odometry or incremental scan matching, and (ii)~loop closure detection, which is needed for data association on a global scale. Loop closures enable SLAM approaches to correct accumulated drift resulting in a globally consistent map.

\begin{figure}
  \centering
  \includegraphics[width=0.72\linewidth]{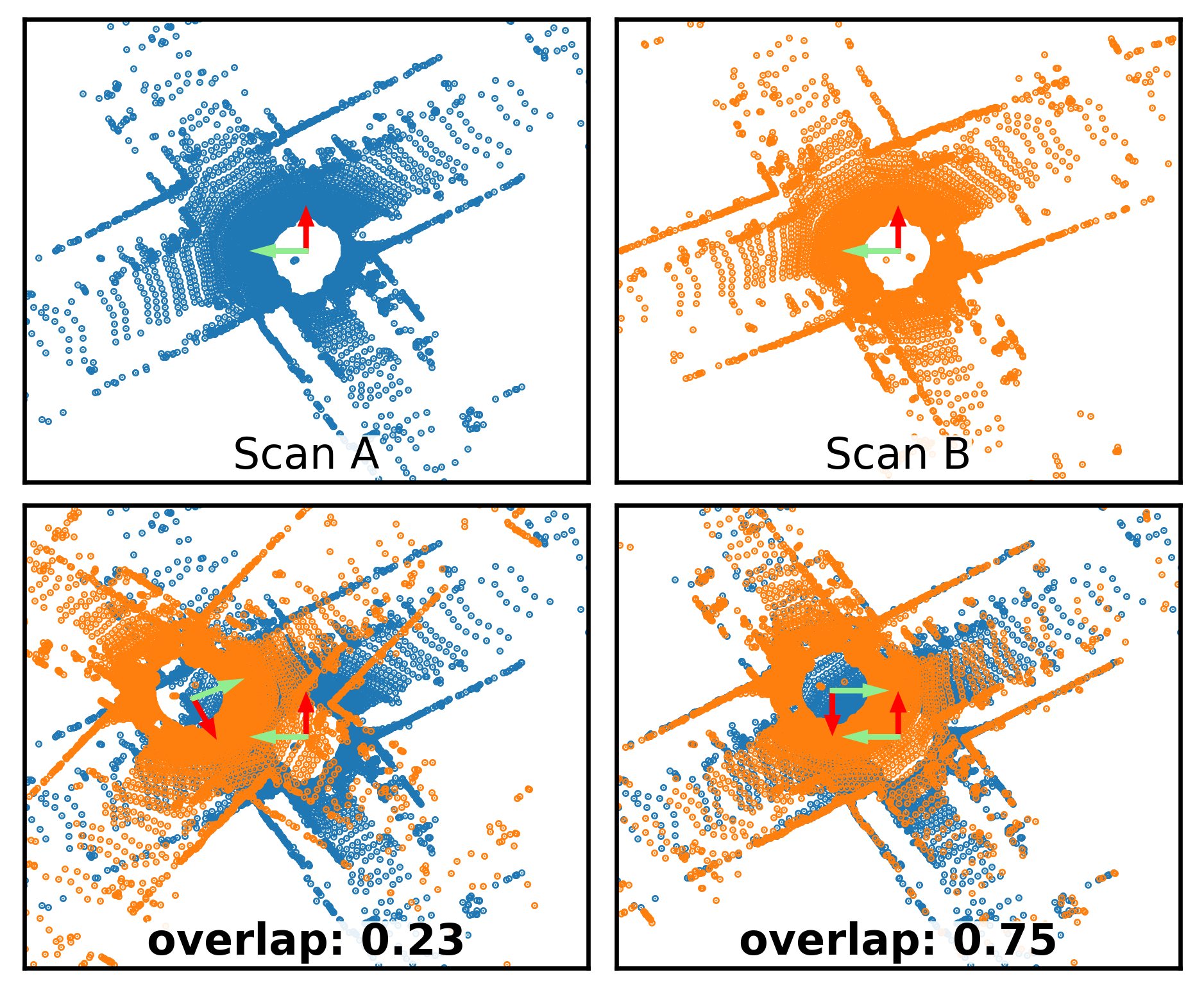}
  \caption{Overlap of two scans (blue and orange points) at a loop closure location but computed with different relative transformations.
    The overlap depends on the relative transformation and larger overlap values often correspond to better alignment between the point clouds.
    Our approach can predict the overlap 
    \textit{without} knowing the relative transformation between the scans.}
  \label{fig:overlap}
   \vspace{-0.2cm}
\end{figure}

In this paper, we propose a new method to loop closing for laser range scans produced by a rotating 3D LiDAR sensor installed on a wheeled robot or similar vehicle.
Instead of using handcrafted features~\cite{he2016iros,steder2011iros}, we propose a deep neural network designed to find loop closure candidates. Our network predicts both, a so-called overlap defined on range images and a relative yaw angle between two 3D LiDAR scans recorded with a typical sensor setup often used on automated cars.
The concept of overlap has been used in photogrammetry to estimate image overlaps, see also~\secref{subsec:overlap_concept} and~\ref{subsec:overlap_computation}, and we use it on LiDAR range images. It is a  useful tool for loop closure detection as illustrated in~\figref{fig:overlap} and can quantify the quality of matches.
The yaw estimate serves as an initial guess for a subsequent application of iterative closest point~(ICP)~\cite{besl1992pami} to determine the relative pose between scans to derive loop closures constraints for the pose graph optimization.
Instead of ICP, one could also use global scan matching~\cite{censi2009icra, zhou2016eccv, Yang2020arxiv} to estimate the relative pose between scans.

The main contribution of this paper is a deep neural network that exploits different types of information generated from LiDAR scans to provide overlap and relative yaw angle estimates between pairs of 3D scans. This information includes depth, normals, and intensity or remission values. We additionally exploit a probability distribution over semantic classes that can be computed for each laser beam.
Our approach relies on a spherical projection of LiDAR scans, rather than the raw point clouds, which makes the proposed OverlapNet comparably lightweight.
We furthermore integrate it into a state-of-the-art SLAM system~\cite{behley2018rss} for loop closure detection and evaluate its performance also with respect to generalization to different environments.

We train the proposed OverlapNet on parts of the KITTI odometry dataset and evaluate it on unseen data.
We thoroughly evaluate our approach, provide ablation studies using different modalities, and test the integrated SLAM system in an online manner.
Furthermore, we provide results for the Ford campus dataset, which was recorded using a different sensor setup in a different country and a differently structured environment.
The experimental results suggest that our method outperforms other state-of-the-art baseline methods and is also able to generalize well to unseen environments.

In sum, our approach is able to
(i)~predict the overlap and relative yaw angle between pairs of LiDAR scans by exploiting multiple cues without using relative poses,
(ii)~combine odometry information with overlap predictions to detect correct loop closure candidates,
(iii)~improve the overall pose estimation results in a state-of-the-art SLAM system yielding more globally consistent maps,
(iv)~solve loop closure detection without prior pose information,
(v)~initialize ICP using the OverlapNet predictions yielding correct scan matching results.
The implementation of our approach is available at: 
\url{https://github.com/PRBonn/OverlapNet}

\section{Related Work}
\label{sec:related}

Loop closure detection using various sensor modalities~\cite{hess2016icra, stachniss2005advancedrobotics} is a classical topic in robot mapping. 
We refer to the article by Lowry \etal~\cite{lowry2016tro} for an overview of approaches using cameras.
Here, we mainly concentrate on related work addressing 3D LiDAR-based approaches.

Steder \etal~\cite{steder2011iros} propose a place recognition system operating on range images generated from 3D LiDAR data that uses a combination of bag-of-words and a NARF-feature-based~\cite{steder2010irosws} relative poses estimation exploiting ideas of FABMAP~\cite{newman2009rss}.
R\"ohling \etal~\cite{roehling2015iros} present an efficient method for detecting loop closures through the use of similarity measures on histograms extracted from 3D LiDAR scans.
The work by He \etal~\cite{he2016iros} presents M2DP, which projects a LiDAR scan into multiple reference planes to generate a descriptor using a density signature of points in each plane.
Besides using pure geometric information, there is also work~\cite{cop2018icra, guo2019ral} exploiting the remission information, \ie, how well LiDAR beams are reflected by a surface, to create descriptors for localization and loop closure detection with 3D LiDAR data.

Motivated by the success of deep learning in computer vision~\cite{krizhevsky2012cacm}, deep learning-based methods have been proposed recently.
Barsan \etal~\cite{barsan2018CoRL} propose a deep network-based localization method, which embeds LiDAR sweeps and intensity maps into a joint embedding space and achieves localization by matching between these embeddings.
Dub\'{e} \etal~\cite{dube2017icra} advocate the usage of segments for loop closure detection.
Cramariuc \etal~\cite{cramariuc2018arxiv} train a CNN to extract descriptors from segments and use it to retrieve near-by place candidates.
Schaupp \etal~\cite{schaupp2019iros} propose a system called OREOS for place recognition, that also estimates the yaw discrepancy between scans.
Furthermore, Yin \etal~\cite{yin2019tits} develop LocNet, which uses semi-handcrafted feature learning based on a siamese network to solve place recognition.
Lu \etal~\cite{Lu2019cvpr} proposed~$L^3$-net, which uses 3D convolutions and a recurrent neural network to learn local descriptors for global localization.
Uy and Lee~\cite{angelina2018cvpr} proposed PointNetVLAD to generate a global descriptor for 3D point clouds.
Kim \etal~\cite{kim2019ral} proposed a learning-based descriptor called SCI to solve long-term global localization.
Most recently, Sun \etal~\cite{sun2020icra} also proposed a learning-based
method combined with Monte Carlo localization to achieve a fast global localization.  

Contrary to the above-mentioned methods, our method exploits multiple types of information extracted from 3D LiDAR scans, including depth, normal information, intensity/remission and probabilities of semantic classes generated by a semantic segmentation system~\cite{milioto2019iros}.

Similar to LocNet~\cite{yin2019tits}, we also use a siamese network, but we learn features and yield predictions end-to-end.
Our network can directly provide estimates for overlap and the relative yaw angle between pairs of LiDAR scans.
Different from OREOS~\cite{schaupp2019iros}, our method not only provides loop closures candidates but also an estimate of the matching quality in terms of the overlap.

Recently, Zaganidis \etal~\cite{zaganidis2019iros} proposed a Normal Distributions Transform (NDT) histogram-based loop closure detection method, which is also assisted by semantic information. In contrast to ours, their method needs a dense global map and cannot estimate the relative yaw angle.

\section{Our Approach}
\label{sec:overlap_network}

\subsection{The Concept of Overlap}
\label{subsec:overlap_concept}

The idea of overlap that we are using here has its origin in the photogrammetry and computer vision community~\cite{hussain2004}.
To successfully match two images and calculate their relative pose, the images must overlap.
This can be quantified by defining the overlap percentage as the percentage of pixels
in the first image, which can successfully be projected back into the second
image without occlusion. Note that this measure is not symmetric: If there is a large scale difference of the image pair, \eg, one image shows a wall and the other shows many buildings around that wall, the overlap percentage for the first to the second image can be large and from the second to the first image low.
In this paper, we use the idea of overlap for range images, exploiting the range information explicitly.

For loop closing, a threshold on the overlap percentage can be used to
decide whether two LiDAR scans are at the same place and/or a loop closing can be done.
For loop closing, this measure maybe even better than the commonly used distance between the recorded positions of a pair of scans, since the positions might be affected by drift and therefore unreliable.
The overlap predictions are independent of the relative poses and can be therefore used to find loop closures without knowing the correct relative pose between scans.
\figref{fig:overlap} shows the overlap of two scans as an example.

\subsection{Definition of the Overlap between Pairs of LiDAR Scans}
\label{subsec:overlap_computation}

We use spherical projections of LiDAR scans as input data, which is often used to speed up computations~\cite{behley2018rss, bogoslavskyi2016iros, chen2019iros}.
We project the point cloud~$\set{P}$ to a so-called vertex map~$\set{V}: \RR^2 \mapsto \RR^3$, where each pixel is mapped to the nearest 3D point.
Each point~$\vec{p}_i = (x, y, z)$ is converted via the function~$\Pi:  \RR^3 \mapsto \RR^2$ to spherical coordinates and finally to image coordinates~$(u,v)$ by
\begin{align}
  \left( \begin{array}{c} u \vspace{0.0em}\\ v \end{array}\right) & = \left(\begin{array}{cc} \frac{1}{2}\left[1-\arctan(y, x) \pi^{-1}\right] w   \vspace{0.5em} \\
      \left[1 - \left(\arcsin(z r^{-1}) + \mathrm{f}_{\mathrm{up}}\right) \mathrm{f}^{-1}\right] h\end{array} \right), \label{eq:projection}
\end{align}
where~$r = ||\vec{p}||_2$ is the range,~$\mathrm{f} = \mathrm{f}_{\mathrm{up}} + \mathrm{f}_{\mathrm{down}}$ is the vertical field-of-view of the sensor, and~$w, h$ are the width and height of the resulting vertex map~$\set{V}$.

For a pair of LiDAR scans~$\set{P}_1$ and~$\set{P}_2$, we generate the corresponding vertex maps~$\set{V}_{1}$,~$\set{V}_{2}$.
We denote the sensor-centered coordinate frame at time step~$t$ as~$C_t$. Each pixel in coordinate frame~$C_t$ is associated with the world frame~$W$ by a pose~$\mat{T}_{WC_t} \in \RR^{4 \times 4}$.
Given the poses~$\mat{T}_{WC_{1}}$ and~$\mat{T}_{WC_{2}}$, we can reproject  scan~$\set{P}_1$ into the coordinate frame of the other's vertex map~$\set{V}_{2}$ and generate a reprojected vertex map~$\set{V}'_{1}$:
\begin{align}
  \set{V}'_{1} & = \Pi \left( \mat{T}_{WC_{1}}^{-1} \mat{T}_{WC_{2}} \set{P}_1 \right).
\end{align}

We then calculate the absolute difference of all corresponding pixels in~$\set{V}'_{1}$ and~$\set{V}_{2}$, considering only those pixels that correspond to valid range readings in both range images. The overlap is then calculated as the percentage of all differences in a certain distance $\epsilon$ relative to all valid entries, \ie, the overlap of two LiDAR scans~$O_{C_1 C_2}$ is defined as follows:
\begin{align}
  O_{C_1 C_2} & = \frac{\sum_{(u,v)} \mathbb{I}\Big\{\left|\left| \set{V}'_{1}(u,v) - \set{V}_{2}(u,v) \right|\right| \leq \epsilon \Big\}} {\min\left( \textrm{valid}(\set{V}'_{1}), \textrm{valid}(\set{V}_{2}) \right)},  \label{eq:overlap}
\end{align}
where~$\mathbb{I}\{a\} = 1$ if~$a$ is true and~$0$ otherwise.~$\textrm{valid}(\set{V})$ is the the number of valid pixels in~$\set{V}$, since not all pixel might have a valid LiDAR measurement associated after the projection.

\begin{figure}[t]
  \vspace{0.2cm}
  \centering
  \subfigure[Exhaustive \newline evaluation of Eq.~(\ref{eq:overlap})]{\includegraphics[width=0.29\linewidth]{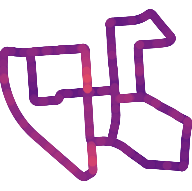}}
  \subfigure[OverlapNet \newline estimates]{\includegraphics[width=0.29\linewidth]{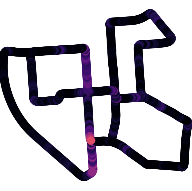}}
  \subfigure[Ground truth]{\includegraphics[width=0.29\linewidth]{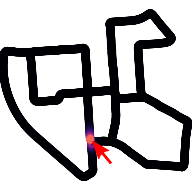}}
  \subfigure{\includegraphics[width=0.08\linewidth]{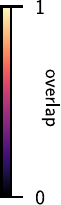}}

  \caption{Overlap estimations of one frame to all others. The red arrow points out the position of the query scan. If we directly use Eq.~(\ref{eq:overlap}) to estimate the overlap between two LiDAR scans without knowing the accurate relative poses, it is hard to decide which pairs of scans are true loop closures, since most evaluations of Eq.~(\ref{eq:overlap}) show high values. In contrast, our OverlapNet can predict  the overlaps between two LiDAR scans well.}
  \vspace{-0.2cm}
  \label{fig:overlap_eq3}
\end{figure}

We use \eqref{eq:overlap} only for creating training data, i.e., only positive examples of correct loop closures get a non-zero overlap assigned using the relative poses between scans, as shown in~\figref{fig:overlap_eq3}(c).
  However, when performing loop closure detection for online SLAM, the approximate relative poses from SLAM before loop closure are not accurate enough to calculate usable overlaps by using~\eqref{eq:overlap} because of accumulated drift.
  We tried directly estimating overlaps using~\eqref{eq:overlap} assuming the relative pose as identity and applying different orientations, \eg, every~$30$ degrees rotation around the vertical axis, and using the maximum over all these overlaps as an estimate. 
  \figref{fig:overlap_eq3} shows the estimated overlaps for all scans using a query scan produced by this method and the result of the estimated overlap for all scans using OverlapNet. We leave out the 100 most recent scans because they will not be loop closure candidates.
  In the case of the exhaustive approach, many wrong loop closure candidates get high overlap values, while our approach performs better since it produces a highly distinctive peak around the correct location.
  Furthermore, it takes on average~$1.2$\,s to calculate the overlap for one pair of scans using the exhaustive approach, which makes it unusable in real-world scenarios. In contrast, the complete OverlapNet needs on average~$17$\,ms for one pair overlap estimation when using depth and normal information only.

\subsection{Overlap Network Architecture}

\begin{figure*}[!t]
  \vspace{0.2cm}
  \centering
  \includegraphics[width=0.85\linewidth]{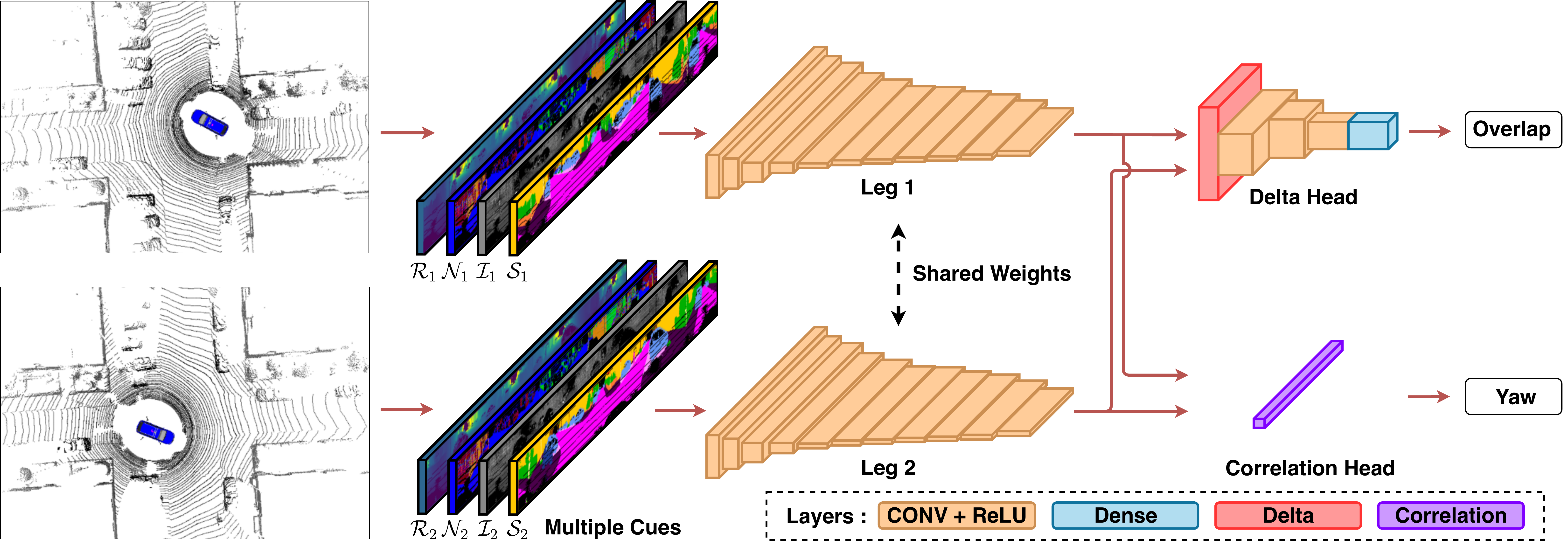}
  \caption{Pipeline overview of our proposed approach.
    The left-hand side shows the preprocessing of the input data which exploits multiple cues generated from a single LiDAR scan, including range~$\set{R}$, normal~$\set{N}$, intensity~$\set{I}$, and semantic class probability~$\set{S}$ information.
    The right-hand side shows the proposed OverlapNet which consists of two legs sharing weights and the two heads use the same pair of feature volumes generated by the two legs.
    The outputs are the overlap and relative yaw angle between two LiDAR scans.}
  \vspace{-0.2cm}
  \label{fig:pipeline}
\end{figure*}

The overview of the proposed OverlapNet is depicted in \figref{fig:pipeline}.
We exploit multiple cues, which can be generated from a single LiDAR scan, including depth, normal,  intensity, and semantic class probability information.
The depth information is stored in the range map~$\set{R}$, which consists of one channel.
We use neighborhood information of the vertex map to generate a normal map~$\set{N}$, which has three channels encoding the normal coordinates.
We directly obtain the intensity information, also called remission, from the sensor and represent the intensity information as a one-channel intensity map~$\set{I}$.
The point-wise semantic class probabilities are computed using RangeNet++~\cite{milioto2019iros} and we represent them as a semantic map~$\set{S}$.
RangeNet++ delivers probabilities for~$20$ different classes. For efficiency sake, we reduce the~$20$-dimensional RangeNet++ output to a compressed $3$-dimensional vector using principal component analysis.
The information is combined into an input tensor of size~$64 \times 900 \times D$, where~$64, 900$ are the height and width of the inputs, and $D$ depends on the types of data used.

Our proposed OverlapNet is a siamese network architecture~\cite{bromley1994ijprai}, which consists of two legs sharing weights and two heads that use the same pair of feature volumes generated by the two legs.
The trainable layers are listed in~\tabref{tab:leg_architecture}.

\subsubsection{Legs}
The proposed OverlapNet has two legs, which have the same architecture and share the same weights.
Each leg is a fully convolutional network (FCN) consisting of~$11$ convolutional layers.
This architecture is quite lightweight and generates feature volumes of size~$1 \times 360 \times 128$.
Note that our range images are cyclic projections and that a change in the yaw angle of the vehicle results in a cyclic column shift of the range image. Thus, the single row in the feature volume can represent a relative yaw angle estimate (because a yaw angle rotation results in a pure horizontal shift of the input maps).
As the FCN is translation-equivariant, the feature volume will be shifted horizontally. The number of columns of the feature volume defines the resolution of the yaw estimation, which is~$1$ degree in the case of our leg architecture.

\begin{table}[t]
  \centering
  \vspace{0.2cm}
  \caption{Layers of our network architecture}
  \scriptsize{
    \begin{tabular}{cccccc}
      \toprule
       & Operator & Stride  & Filters & Size    & Output Shape               \\
      \midrule
      \multirow{11}{*}{\begin{sideways}Legs\end{sideways}}
       & Conv2D   & (2, 2)  & 16     & (5, 15) & ~$30 \times 443 \times 16$ \\
       & Conv2D   & (2, 1)  & 32     & (3, 15) & ~$14 \times 429 \times 32$ \\
       & Conv2D   & (2, 1)  & 64     & (3, 15) & ~$6 \times 415 \times 64$  \\
       & Conv2D   & (2, 1)  & 64     & (3, 12) & ~$2 \times 404 \times 64$  \\
       & Conv2D   & (2, 1)  & 128    & (2, 9)  & ~$1 \times 396 \times 128$ \\
       & Conv2D   & (1, 1)  & 128    & (1, 9)  & ~$1 \times 388 \times 128$ \\
       & Conv2D   & (1, 1)  & 128    & (1, 9)  & ~$1 \times 380 \times 128$ \\
       & Conv2D   & (1, 1)  & 128    & (1, 9)  & ~$1 \times 372 \times 128$ \\
       & Conv2D   & (1, 1)  & 128    & (1, 7)  & ~$1 \times 366 \times 128$ \\
       & Conv2D   & (1, 1)  & 128    & (1, 5)  & ~$1 \times 362 \times 128$ \\
       & Conv2D   & (1, 1)  & 128    & (1, 3)  & ~$1 \times 360 \times 128$ \\
      \midrule
      \multirow{3}{*}{\begin{sideways}Delta Head\end{sideways}}
       & Conv2D   & (1, 15) & 64     & (1, 15) & ~$360 \times 24 \times 64$ \\
       & Conv2D   & (15, 1) & 128    & (15, 1) & ~$24 \times 24 \times 128$ \\
       & Conv2D   & (1, 1)  & 256    & (3, 3)  & ~$22 \times 22 \times 256$ \\
       & Dense    & -       & -      & -       & ~$1$                       \\
      \bottomrule
    \end{tabular}
  }
  \label{tab:leg_architecture}
\end{table}

\subsubsection{Delta Head}

The delta head is designed to estimate the overlap between two scans. 
It consists of a delta layer, three convolutional layers, and one fully connected layer.

\begin{figure}[t]
  \vspace{0.2cm}
  \centering
  \includegraphics[width=0.7\linewidth]{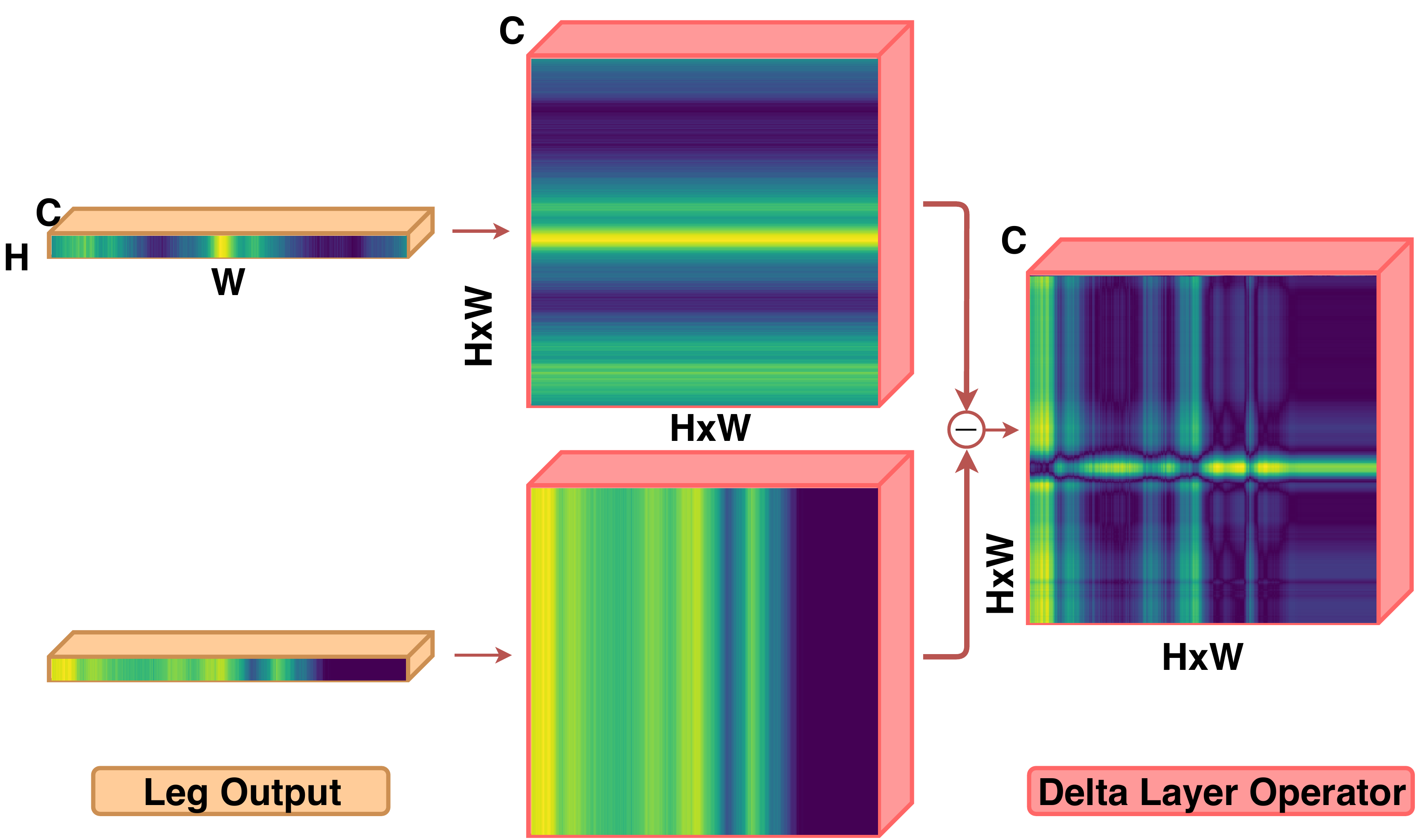}
  \caption{Delta layer.
    Computation of pairwise differences is efficiently performed by concatenating the feature volumes and transposition of one concatenated feature volume.}
  \vspace{-0.2cm}
  \label{fig:delta_layer}
\end{figure}

The delta layer, shown in \figref{fig:delta_layer}, computes all possible absolute differences of all pixels.
It takes the output feature volumes~$\mat{L}^l \in \RR^{H\times W\times C}$ from the two legs~$l$ as input.
These are stacked in a tiled tensor $\mat{T}^l \in \RR^{HW \times HW \times C}$ as follows:
\begin{align}
\mat{T}^0(iW+j, k, c) &= \mat{L}^0(i, j, c) \\ 
\mat{T}^1(k, iW+j, c) &= \mat{L}^1(i, j, c),
\end{align} 
with $k= \{0, \dots, HW-1\}$, $i= \{0, \dots, H-1\}$ and \mbox{$j = \{0,\dots, W-1\}$}. 

Note that~$\mat{T}^1$ is transposed in respect to~$\mat{T}^0$, as depicted in the middle of~\figref{fig:delta_layer}.
After that, all differences are calculated by element-wise absolute differences between~$\mat{T}^0$ and~$\mat{T}^1$.

By using the delta layer, we can obtain a representation of the latent difference information,
which can be later exploited by the convolutional and fully-connected layers to estimate the overlap. 
Different overlaps induce different patterns in the output of the delta layer.

\subsubsection{Correlation Head}

The correlation head~\cite{nagashima2007iciar} is designed to estimate the yaw angle between two scans using the feature volumes of the two legs.
To perform the cross-correlation, we first pad horizontally one feature volume by copying the same values (as the range images are cyclic projections around the yaw angle). This doubles the size of the feature volume. We then use the other feature volume as a kernel that is shifted over the first feature volume generating a 1D output of size~$360$. The~$\argmax$ of this feature serves as the estimate of the relative yaw angle of the two input scans with a~$1$ degree resolution.

\subsection{Loss Functions}
We train our OverlapNet end-to-end to estimate the overlap and the relative yaw angle between two LiDAR scans at the same time.
Typically, to train a neural network one needs a large amount of manually labeled ground truth data. 
In our case, this is~($I_1$, $I_2$, $Y_{O}$, $Y_{Y}$), where~$I_1, I_2$ are two inputs and~$Y_{O}, Y_{Y}$ are the ground truth overlaps and the ground truth yaw angles respectively.
We are however able to generate the input and the ground truth without any manual effort in a fully automated fashion given a dataset with pose information.
From given poses, we can calculate the ground truth overlap and relative yaw angles directly.
We denote the legs part network with trainable weights as~$f_{L}(\cdot)$, the delta head as~$f_{D}(\cdot)$ and the correlation head as~$f_{C}(\cdot)$.

For training, we combine the loss~$L_O(\cdot)$ for the overlap and the loss~$L_Y(\cdot)$  for the yaw angle using a weight~$\alpha$:
\begin{align}
  L\left( I_1, I_2, Y_{O}, Y_{Y} \right)\hspace{-0.10em} & =\hspace{-0.10em}
  L_O\left( I_1, I_2, Y_{O} \right)\hspace{-0.2em}+\hspace{-0.2em}\alpha L_Y\hspace{-0.1em}\left( I_1, I_2, Y_{Y} \right). \label{eq:combined_loss}
\end{align}

We treat the overlap estimation as a regression problem and use a weighted absolute difference of ground truth~$Y_{O}$ and network output~$\hat Y_{O} = f_{D} \left( f_{L}\left( I_1 \right), f_{L}\left( I_2 \right) \right)$ as the loss function. For weighting, we use a scaled sigmoid function:
\begin{align}
  L_O\left( I_1, I_2, Y_{O} \right) & = \textrm{sigmoid}\left(s \left(\left|\hat Y_{O} - Y_{O} \right| + a \right) - b \right), \label{eq:overlap_loss}
\end{align}
with~$\textrm{sigmoid}(v) = (1+\exp(-v))^{-1}$, $a, b$ are offsets and~$s$ being a scaling factor.

For the yaw angle estimation, we use a lightweight representation of the correlation head output, which leads to a one-dimensional vector of size~$360$.
We take the index of the maximum, the~$\argmax$, as the estimate of the relative angle in degrees.
As the~$\argmax$ is not differentiable, we cannot treat this as a simple regression problem.
The yaw angle estimation, however, can be regarded as a binary classification problem that decides for every entry of the head output whether it is the correct angle or not.
Therefore, we use the binary cross-entropy loss given by
\begin{align}
   & L_Y\left( I_1, I_2, Y_{Y} \right)=\sum\nolimits_{i=\{1, \dots, N\}} \textrm{H}\left(Y_{Y}^i,  \hat Y_{Y}^i\right),
\end{align}
where~$\textrm{H}(p,q) = p\log(q) - (1-p)\log(1-q)$ is the binary cross entropy and~$N$ is the size of the output 1D vector. \mbox{$\hat Y_{Y} = f_{C} \left( f_{L}\left( I_1 \right), f_{L}\left( I_2 \right) \right)$} is the relative yaw angle estimate. Note that we only train the network to estimate the relative yaw angle of a pair of scans with overlap larger than~$30\%$, since this minimum overlap was needed to result in correct pose estimates of the ICP as explained in \secref{subsec:lcd}, but also experimentally validated in \secref{sec:icp_initial_guess}.

\subsection{SLAM Pipeline}

We use the surfel-based mapping system called SuMa~\cite{behley2018rss} as our SLAM pipeline and integrate OverlapNet in SuMa replacing its original heuristic loop closure detection method.
We only summarize here the steps of SuMa relevant to our approach and refer for more details to the original paper~\cite{behley2018rss}.

SuMa uses the same vertex map~$\set{V}_D$ and normal map~$\set{N}_D$ as discussed  in \secref{subsec:overlap_computation}.
Furthermore, SuMa uses projective ICP with respect to a rendered map view~$\set{V}_M$ and~$\set{N}_M$ at timestep~$t-1$, the pose update~$\mat{T}_{C_{t-1}C_{t}}$ and consequently~$\mathbf{T}_{WC_t}$ by chaining all pose increments.
Therefore, each vertex~$\vec{u} \in \set{V}_D$ is projectively associated to a reference vertex~$\vec{v}_\vec{u} \in \set{V}_M$.
Given this association information, SuMa estimates the transformation between scans by incrementally minimizing the point-to-plane error given by
\begin{align}
  E(\set{V}_D, \set{V}_M, \set{N}_M) & = \sum_{\vec{u} \in \set{V}_D} \bigg(\vec{n}_{u}^\top \left(\mat{T}^{(k)}_{C_{t-1}C_{t}}\vec{u} - \vec{v}_{u}\right)\bigg)^2. \label{eq:objective}
\end{align}

Each vertex~$\vec{u} \in \set{V}_D$ is projectively associated to a reference vertex~$\vec{v}_{u} \in \set{V}_M$ and its  normal~$\vec{n}_{u} \in \set{N}_M$ via
\begin{align}
  \vec{v}_{u} & = \set{V}_M\left(\Pi\left(\mat{T}^{(k)}_{C_{t-1}C_{t}}\vec{u}\right)\right)  \\
  \vec{n}_{u} & = \set{N}_M\left(\Pi\left(\mat{T}^{(k)}_{C_{t-1}C_{t}}\vec{u}\right)\right).
\end{align}

SuMa then minimizes the objective of \eqref{eq:objective} using Gauss-Newton and determines increments~$\delta$ by iteratively solving
\begin{align}
  \delta & = \left(\mat{J}_\delta^\top \mat{W} \mat{J}_\delta\right)^{-1}\mat{J}_\delta^\top\mat{W}\vec{r},
\end{align}
where~$\mat{W} \in \RR^{n\times n}$ is a diagonal matrix containing weights~$w_u$,~$\vec{r} \in \RR^n$ is the stacked residual vector, and~$\mat{J}_\delta \in \RR^{n\times 6}$ the Jacobian of~$\vec{r}$ with respect to the increment~$\delta$.

SuMa employs a loop closure detection module, which considers the nearest frame in the built map as the candidate for loop closure given the current pose estimate.
Loop closure detection works well for small loops, but the heuristic fails in areas with only a few large loops.
Furthermore, drift in the odometry estimate can lead to large displacements, where the heuristic of just taking the nearest frame in the already mapped areas does not yield correct candidates, which will be also shown in our experiments.

\subsection{Covariance Propagation for Geometric Verification}
\label{sec:cov_prop}

SuMa's loop closure detection uses a fixed search radius.
In contrast, we use the covariance of the pose estimate and error propagation to automatically adjust the search radius.

We assume a noisy pose~$\mathbf{T}_{C_{t-1}C_{t}} = \{ \bar{\mathbf{T}}_{C_{t-1}C_{t}}, \Sigma_{C_{t-1}C_{t}} \}$ with mean~$\bar{\mathbf{T}}_{C_{t-1}C_{t}}$ and covariance~$\Sigma_{C_{t-1}C_{t}}$.
We can estimate the covariance matrix  by
\begin{align}
  \Sigma_{C_{t-1}C_{t}} & = \frac{1}{K} \frac{E}{N-M}\left(\mat{J}_\delta^\top \mat{W} \mat{J}_\delta\right)^{-1},
\end{align}
where~$K$ is the correction factors of the Huber robustized covariance estimation~\cite{huber1981book},~$E$ is the sum of the squared point-to-plane errors (sum of squared residuals) given the pose~$\mathbf{T}_{C_{t-1}C_{t}}$, see \eqref{eq:objective},~$N$ is the number of correspondences,~$M=6$ is the dimension of the transformation between two 3D poses.

To estimate the propagated uncertainty during the incrementally pose estimation, we can update the mean and covariance as follows:
\begin{align}
  \bar{\mat{T}}_{C_{t-1}C_{t+1}} & = \bar{\mat{T}}_{C_{t-1}C_{t}} \bar{\mat{T}}_{C_{t}C_{t+1}}                                              \\
  \Sigma_{C_{t-1}C_{t+1}}        & \approx \Sigma_{C_{t-1}C_{t}}+ \mat{J}_{C_{t}C_{t+1}}^\top \Sigma_{C_{t}C_{t+1}} \mat{J}_{C_{t}C_{t+1}},
\end{align}
where~$\mat{J}_{C_{t}C_{t+1}}$ is the Jacobian of~${\mat{T}}_{C_{t}C_{t+1}}$. 

Since we need the Mahalanobis distance~$D_M$ as a probabilistic distance measure between two poses, we make use of Lie algebra to express $\mat{T}$ as a 6D vector~$\vec{\xi} \in \mathfrak{se}(3)$ using $\vec{\xi} = \log{\mat{T}}$,
yielding
\begin{align}
  D_M\left(\mat{T}_{C1}, \mat{T}_{C2}\right) & = \sqrt{\Delta\vec{\xi}_{C1C2}^\top \Sigma_{C1C2}^{-1} \Delta \vec{\xi}_{C1C2}}.
\end{align}

Using the scaled distance, we can now restrict the search space depending on the pose uncertainty to save computation time. 
However, we can use our framework also without any prior information, \ie, perform place recognition.

\section{Experimental Evaluation}
\label{sec:exp}
The experimental evaluation is designed to support the key claims that our approach is able to:
(i)~predict the overlap and relative yaw angle between pairs of LiDAR scans by exploiting multiple cues without given poses,
(ii)~combine odometry information with overlap predictions to detect correct loop closure candidates,
(iii)~improve the overall pose estimation results in graph-based SLAM yielding more globally consistent maps,
(iv)~solve loop closure detection without prior pose information,
(v)~initialize ICP using OverlapNet predictions yielding correct scan matching results.

We train and evaluate our approach on the KITTI odometry benchmark~\cite{geiger2012cvpr},
which provides LiDAR scans recorded with a Velodyne HDL-64E of urban areas around Karlsruhe in Germany.
We follow the experimental setup of Schaupp \etal~\cite{schaupp2019iros} and use sequence~$00$ for evaluation.
Sequences~$03-10$ are used for training and sequence~$02$ is used for validation.

To evaluate the generalization ability of our method, we also test it on the Ford campus dataset~\cite{pandey2011ijrr}, which is recorded on the Ford research campus and downtown Dearborn in Michigan using a different version of the Velodyne HDL-64E.
In the case of the Ford campus dataset, we test our method on sequence~$00$ which has several large loops.
Note that we never trained our approach on the Ford campus dataset.

For generating overlap ground truth, we only use points within a distance of~$75$\,m to the sensor.
For overlap computation, see \eqref{eq:overlap}, we use~$\epsilon = 1\,$m.
We use a learning rate of~$10^{-3}$ with a decay of~$0.99$ every epoch and train at most~$100$ epochs.
For the combined loss, \eqref{eq:combined_loss}, we set~$\alpha = 5$.
For the overlap loss, \eqref{eq:overlap_loss}, we use~$a = 0.25, b = 12,$ and scale factor~$s = 24$.

\subsection{Loop Closure Detection}
\label{subsec:lcd}

\begin{figure}[t]
  \vspace{0.2cm}
  \centering
  \includegraphics[]{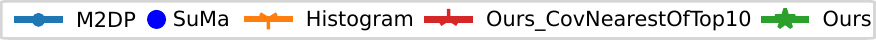}
  \subfigure[KITTI sequence~$00$]{\includegraphics[]{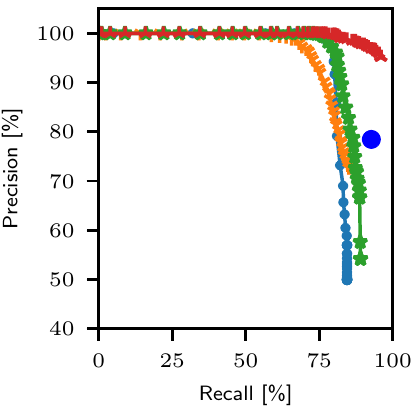}}
  \subfigure[Ford campus sequence~$00$]{\includegraphics[]{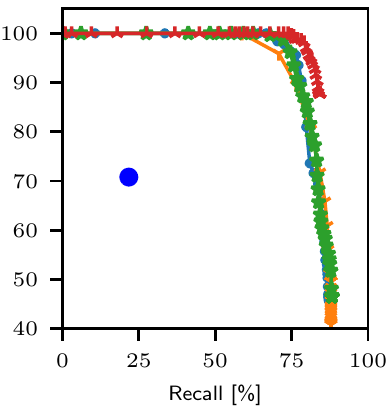}}
  \caption{Precision-Recall curves of different approaches.}
  \label{fig:lcd_prc}
  \vspace{-0.2cm}
\end{figure}

In our first experiments, we investigate the loop closure performance of our approach and compare it to existing methods.
Loop closure detection typically assumes that robots revisit places during the mapping while moving with uncertain odometry.  
Therefore, the prior information about robot poses extracted from the pose graph is available for the loop closure detection.
The following criteria are used in these experiments: 
\begin{itemize}[leftmargin=*]
  \item To avoid detecting a loop closure in the most recent scans, we do not search candidates in the latest~$100$ scans.
  \item For each query scan, only the best candidate is considered throughout this evaluation.
\item
  Most SLAM systems search for potential closures only within the~$3\sigma$ area around the current pose estimate.
We do the same, either using the Euclidean or the Mahalanobis distance, depending on the approach.
\item We use a relatively low threshold of~$30$\,\% for the overlap to decide if a candidate is a true positive.
  We aim to find more loops even in some challenging situations with low overlaps,~\eg, when the car drives back to an intersection from the opposite direction (as highlighted in the supplementary video\footnote{\label{note:video}https://youtu.be/YTfliBco6aw}). 
  Furthermore, ICP can find correct poses 
  if the overlap between pairs of scans is around~$30$\,\%, as illustrated in the experimental evaluation.
\end{itemize}

We evaluate OverlapNet on both KITTI sequence~$00$ and Ford campus sequence~$00$ using the precision-recall curves shown in \figref{fig:lcd_prc}.
We compare our method, trained with two heads and all cues (labeled as \textit{Ours (AllChannel, TwoHeads)}) with three state-of-the-art approaches, M2DP~\cite{he2016iros}, Histogram~\cite{roehling2015iros}, and the original SuMa~\cite{behley2018rss}. Since SuMa always uses the nearest frame as the candidate for loop closure detection, we can only get one pair of precision and recall value resulting in a single point.
We also show the result of our method using prior information, named \textit{Ours\_CovNearestOfTop10}, which uses covariance propagation (\secref{sec:cov_prop}) to define the search space with the Mahalanobis distance and use the nearest in Mahalanobis distance of the top~$10$ predictions of OverlapNet as the loop closure candidates.

\tabref{tab:lcd_com2others} shows the comparison between our approach and the state of the art using the F1 score and the area under the curve (AUC) on both KITTI and Ford campus dataset. For the KITTI dataset, our approach uses the model trained with all cues, including depth, normals, intensity, and a probability distribution over semantic classes. For the Ford campus dataset, our approach uses the model trained with geometric information only, namely \textit{Ours (GeoOnly)}, since other cues are not available in this dataset. We can see that our method outperforms the other methods on the KITTI dataset and attains a similar performance on the Ford campus dataset.
There are two reasons to explain the worse performance on the Ford campus dataset.
First, we never trained our network on the Ford campus dataset or even US roads, and secondly, there is only geometric information available on the Ford campus dataset.
However, our method outperforms all baseline methods in both, KITTI and Ford campus dataset, if we integrate prior information.

We also show the performance in comparison to variants of our method in \tabref{tab:lcd_com2variants}.
We compare our best model \textit{AllChannel} using two heads and all available cues to a variant which only uses a basic multilayer perceptron as the head named \textit{MLPOnly} which consists of two hidden fully connected layers and a final fully connected layer with two neurons (one for overlap, one for yaw angle). The substantial difference of the AUC and F1 scores shows that such a simple network structure is not sufficient to get a good result. Training the network with only one head (only the delta head for overlap estimation, named \textit{DeltaOnly}), has not a significant influence on the performance. 
A huge gain can be observed when regarding the nearest frame in Mahalanobis distance of the top~$10$ candidates in overlap percentage (\textit{CovNearestOfTop10}).

\setlength\tabcolsep{3.5pt}\begin{table}[t]\centering\footnotesize{
\vspace{0.2cm}\caption{Comparison with state of the art.}
\label{tab:lcd_com2others}
\begin{tabular}{l|L{3.5cm}|C{0.9cm}C{0.9cm}}
\toprule 
Dataset & Approach&AUC&F1 score\\
\midrule
\multirow{4}{*}{\parbox{1.8cm}{KITTI}}&Histogram \cite{roehling2015iros}  & 0.83 & 0.83\\
&M2DP \cite{he2016iros}  & 0.83 & 0.87\\
&SuMa \cite{behley2018rss}  & - & 0.85\\
&Ours (AllChannel, TwoHeads) & $\mathbf{ 0.87}$& $\mathbf{ 0.88}$\\
\midrule
\multirow{4}{*}{\parbox{1.8cm}{Ford Campus}}&Histogram \cite{roehling2015iros}  & 0.84 & 0.83\\
&M2DP \cite{he2016iros}  & 0.84& $\mathbf{ 0.85}$\\
&SuMa \cite{behley2018rss}  & - & 0.33\\
&Ours (GeoOnly) & $\mathbf{ 0.85}$ & 0.84\\
\bottomrule
\end{tabular}}
\setlength\tabcolsep{6.0pt}
\vspace{-0.2cm}
\end{table}

\setlength\tabcolsep{3.5pt}\begin{table}[t]\centering\footnotesize{
\vspace{0.2cm}\caption{Comparison with our variants.}
\label{tab:lcd_com2variants}
\begin{tabular}{l|L{3.5cm}|C{0.9cm}C{0.9cm}}
\toprule 
Dataset & Variant&AUC&F1 score\\
\midrule
\multirow{5}{*}{\parbox{1.8cm}{KITTI}}&MLPOnly  & 0.58 & 0.65\\
&DeltaOnly  & 0.85 & 0.88\\
&CovNearestOfTop10 & $\mathbf{ 0.96}$& $\mathbf{ 0.96}$\\
&Ours (AllChannel, TwoHeads)  & 0.87 & 0.88\\
\midrule
\multirow{2}{*}{\parbox{1.8cm}{Ford Campus}}&Ours (GeoOnly)  & 0.85 & 0.84\\
&GeoCovNearestOfTop10 & $\mathbf{ 0.85}$& $\mathbf{ 0.88}$\\
\bottomrule
\end{tabular}}
\setlength\tabcolsep{6.0pt}
\vspace{-0.2cm}
\end{table}

\subsection{Qualitative Results}

\begin{figure}[t]
  \vspace{0.2cm}
  \centering
  \subfigure[SuMa]{\includegraphics[width=0.432\linewidth]{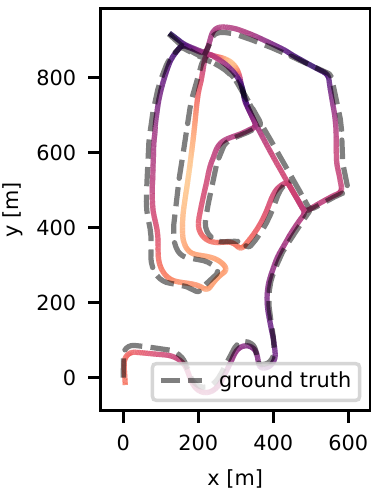}}
  \subfigure[Ours\_CovNearestOfTop10]{
    \includegraphics[width=0.48\linewidth]{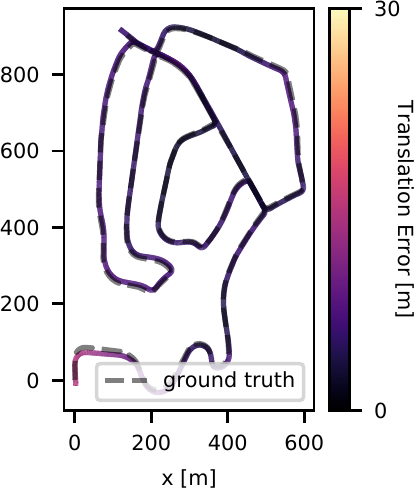}}
  \caption{Qualitative result on KITTI sequence~$02$.}

  \label{fig:Qualitative_Results}
\end{figure}

The second experiment is designed to support the claim that our method is able to improve the overall mapping result.
\figref{fig:Qualitative_Results} shows the odometry results on KITTI sequence 
~$02$. The color in \figref{fig:Qualitative_Results} shows the 3D 
translation error (including height). 
The left figure shows the SuMa and the right figure shows \textit{Ours\_CovNearestOfTop10} using the proposed OverlapNet to detect loop closures. 
We can see that after integrating our method, the overall odometry is much more accurate since we can provide more loop closure candidates with higher accuracy in terms of overlap. The colors represent the translation error of the estimated poses with respect to the ground truth.
Furthermore, after integrating the proposed OverlapNet, the SLAM system can find more loops even in some challenging situations,~\eg, when the car drives back to an intersection from the opposite direction, which is highlighted in the supplementary video\footnoteref{note:video}.

\subsection{Loop Closure Detection without Odometry Information}
\label{subsec:place_recognition}

\begin{figure}[t]
  \centering
  \includegraphics[width=0.9\linewidth]{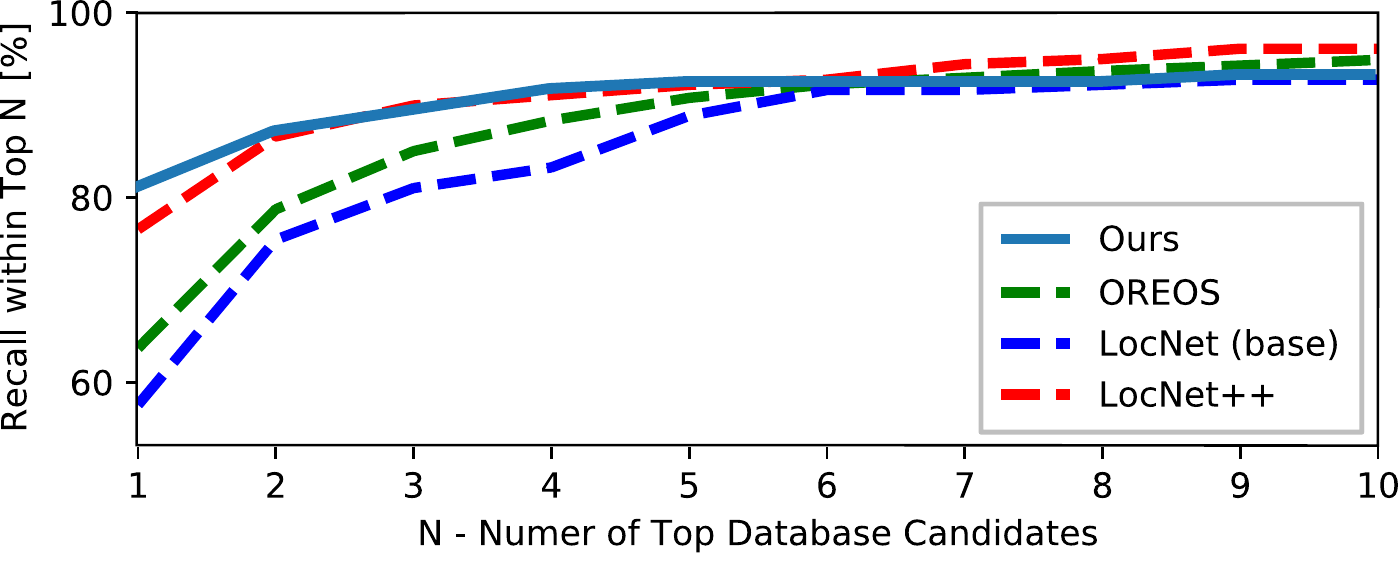}
  \caption{Loop closure detection performance on KITTI sequence~$00$.}
  \label{fig:recall_top_nums}
\end{figure}

The third experiment is designed to support the claim that our approach is well-suited for solving the more general loop closure detection task without using odometry information.

In this case, we assume that we have no prior information about the robot pose.
To compare with the state-of-the-art method OREOS~\cite{schaupp2019iros}, we follow their experimental setup and refer to the original paper for more details.
The OREOS results are those produced by the authors of OREOS.

The respective loop closure candidates recall results are shown in \figref{fig:recall_top_nums}. Our method outperforms all the baseline methods with a small number of candidates and attains similar performance as baseline methods for higher values of numbers of candidates.
However, OREOS and LocNet++ attain a slightly higher recall if more candidates are considered.

\subsection{Yaw Estimation}

\begin{table}[t]
  \vspace{0.2cm}
  \centering
  \caption{Yaw estimation errors without ICP}
  \label{tab:orientation_errors}
  \begin{tabular}{L{3.5cm}C{1.3cm}C{1.2cm}C{1.2cm}}
    \toprule
    Approach     & Mean[deg]        & std[deg]         & Recall[\%]                                                                                    \\
    \midrule
    FPFH+RANSAC* &$13.28$           & $32.19  $          & $97  $                                                                                          \\
    OREOS*       & $12.67$            & $15.23$            & $100 $                                                                                          \\
    Ours (AllChannel, TwoHeads)        & $\mathbf{1.13}$ & $\mathbf{3.34}$ & $\mathbf{100}$                                                                               \\
    \bottomrule
    \multicolumn{4}{p{0.8 \linewidth}}{\scriptsize{\vspace{0.01cm} *: The results are those produced by the authors of OREOS~\cite{schaupp2019iros}.}} \\
  \end{tabular}

\end{table}

\begin{figure}[t]
  \vspace{-3mm}
  \centering
  \subfigure[KITTI sequence~$00$]{\includegraphics[]{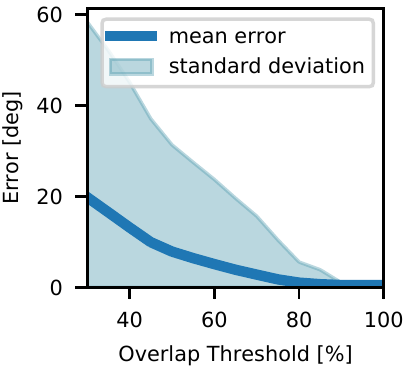}}
  \subfigure[Ford campus sequence~$00$]{\includegraphics[]{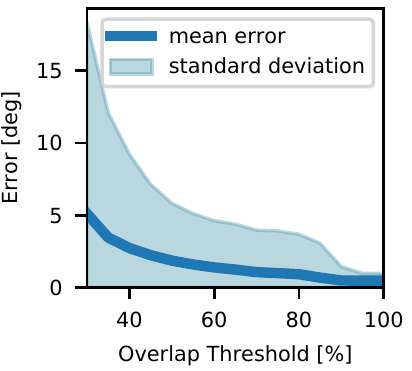}}
  \caption{Overlap and yaw estimation relationship.}
  \label{fig:orientation_plot}
  \vspace{-0.2cm}
\end{figure}

We aim at supporting our claim that our network provides good relative yaw angle estimates.
We use the same experimental setup as described in \secref{subsec:place_recognition}.
\tabref{tab:orientation_errors} summarizes the yaw angle errors
on KITTI sequence~$00$. 

We can see that our method outperforms the other methods in terms of mean error and standard deviations.
In terms of recall, OverlapNet and OREOS always provide a yaw angle estimate, since both approaches are designed to estimate the relative yaw angle for any pairs of scans in contrast to the RANSAC-based method that sometimes fails.

The superior performance can be mainly attributed to the correlation head exploiting the fact that the orientation in LiDAR scans can be well represented by the shift in the range projection.
Therefore, it is easier to train the correlation head to accurately predict the relative yaw angles rather than a multilayer perceptron used in OREOS~\cite{schaupp2019iros}. 
Furthermore, there is also a strong relationship between overlap and yaw angle, which also improves the results when trained together.

\figref{fig:orientation_plot} shows the relationship between real overlap and yaw angle estimation error.
As expected, the yaw angle estimate gets better with increasing overlap.
Based on these plots, our method not only finds candidates but also measures the quality, \ie, when the overlap of two scans is larger than~$90\%$, our method can accurately estimate the relative yaw angle with an average error of about only~$1$ degree.

\subsection{Ablation Study on Input Modalities}

\begin{table}[t]\centering\footnotesize{
\vspace{0.2cm}\caption{Ablation study on usage of input modalities.}
\label{tab:ablation}
\begin{tabular}{C{0.8cm}C{0.98cm}C{1.00cm}C{1.18cm}C{0.6cm}C{0.6cm}C{0.7cm}C{0.7cm}}
\toprule 
\multirow{2}{*}{Depth} & \multirow{2}{*}{Normals} & \multirow{2}{*}{Intensity} & \multirow{2}{*}{Semantics} &\multicolumn{2}{c}{overlap} &\multicolumn{2}{c}{yaw angle[deg]}\\
 & & & &AUC &F1 &Mean &Std\\
\midrule
\cmark &  &  &   & 0.86 & 0.87 & 11.67 & 25.32\\
\cmark & \cmark &  &   & 0.86 & 0.85 & 2.97 & 14.28\\
\cmark & \cmark & \cmark &   & 0.87 & 0.87 & 2.53 & 14.56\\
\cmark & \cmark & \cmark & \cmark  & 0.87 & 0.88 & 1.13 & 3.34\\
\bottomrule
\end{tabular}}
\end{table}

An ablation study on the usage of different inputs is shown in \tabref{tab:ablation}. 
As can be seen, when employing more input modalities, the proposed method is more robust.
We notice that exploiting only depth information with OverlapNet can already perform reasonable in terms of overlap prediction, while it does not perform well in yaw angle estimation. When combining with normal information, the OverlapNet can perform well in both tasks.
Another interesting finding is the drastic reduction of yaw angle mean error and standard deviation when using semantic information.
One reason could be that adding semantic information will make the input data more distinguishable when the car drives in symmetrical environments.
We also notice that semantic information 
will increase the computation time, see \secref{subsec:runtime}.
However, from the ablation study, one could also notice that the proposed method can also achieve good performance by only employing geometric information (depth and normals).

\subsection{Using OverlapNet Predictions as Initial Guesses for ICP}
\label{sec:icp_initial_guess}
\begin{figure}[t]
  \vspace{-3mm}
  \centering
  \subfigure[With identity as initial guess]{\includegraphics[]{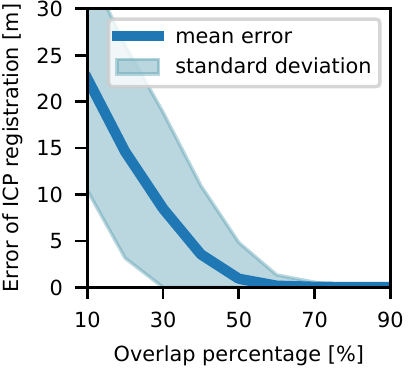}}
  \subfigure[With yaw angle as initial guess]{\includegraphics[]{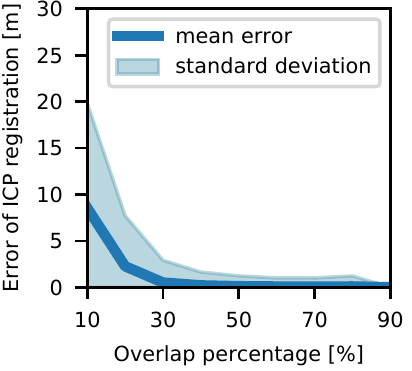}}
  \caption{ICP using OverlapNet predictions as initial guess. The error of ICP registration here is the Euclidean distance between the estimated translation and the ground-truth translation.}
  \label{fig:ICP_overlapnet}
  \vspace{-0.3cm}
\end{figure}

We aim at supporting the claim that our network provides good initializations for ICP with 3D laser scans collected on autonomous cars.
\figref{fig:ICP_overlapnet} shows the relation between the overlap and ICP registration error with and without using OverlapNet predictions as initial guesses. 
The error of ICP registration is here depicted by the Euclidean distance between the estimated relative translation and the ground-truth translation. 
As can be seen, the yaw angle prediction of the OverlapNet increases the chance to get a good result from the ICP even if two frames are relatively far away from each other (with low overlap). 
Therefore in some challenging cases,~\eg the car drives back into an intersection from a different street, our approach can still find loop closures (see in the supplementary video\footnoteref{note:video}).
The results also show that the overlap estimates measure the quality of the found loop closure: larger overlap values result in better registration results of the involved ICP.

\subsection{Runtime}
\label{subsec:runtime}

We tested our method on a system equipped with an Intel i7-8700 with~$3.2$\,GHz and an Nvidia GeForce GTX1080 Ti with~$11$\,GB memory.

For the KITTI sequence~$00$, we could exploit all input cues including the semantic classes provided by RangeNet++~\cite{milioto2019iros}.
We need on average~$75\,$ms per frame for the input data preprocessing,~$6\,$ms per frame for the legs feature extraction,~$27\,$ms per frame for the head matching. 
The worst case for the head matching takes~$630\,$ms for all candidates in the search space.

For the Ford campus dataset, we used only geometric information, which could be generated in~$10\,$ms on average per frame,~$2\,$ms for feature extraction and~$24\,$ms for matching with the worst case of~$550\,$ms.
In real SLAM operation, we only search loop closure candidates inside a certain search space given by pose uncertainty using the Mahalanobis distance (see \secref{sec:cov_prop}). 
Therefore, our method can achieve online operation in long-term tasks, since we usually only have to evaluate a small number of candidate poses.

\section{Conclusion}
\label{sec:conclusion}

In this paper, we presented a novel approach for LiDAR-based loop closure detection. It is based on the overlap between LiDAR scan range images and provides a measure for the quality of the loop closure.
Our approach utilizes a siamese network structure to leverage multiple cues and allows us to estimate the overlap and relative yaw angle between scans.
The experiments on two different datasets suggest that when combined with odometry information our method outperforms other state-of-the-art methods and that it generalizes well to different environments never seen during training.

Despite these encouraging results, there are several avenues for future research. 
First, we want to investigate the integration of other input modalities, such as vision and radar information.
We furthermore plan to test our approach with other datasets collected in different seasons.

\section*{Acknowledgments}
  This work has been supported in part by the German Research Foundation (DFG) under Germany's Excellence Strategy, EXC-2070 - 390732324 (PhenoRob) and under grant number \mbox{BE 5996/1-1} as well as by the Chinese Scholarship Committee.


\bibliographystyle{plainnat}

\bibliography{glorified,new}

\end{document}